\documentclass{sig-alternate}
\usepackage{url}
\usepackage{graphicx}
\usepackage{epstopdf}
\usepackage{multirow}
\usepackage{slashbox}
\usepackage[english]{babel}
\usepackage{verbatim}
\usepackage{scalefnt}
\usepackage{pbox}
\usepackage{array}
\usepackage{subfig}
\usepackage{subfiles}
\usepackage{siunitx}
\usepackage{filecontents}
\usepackage[numbers,sort&compress]{natbib}
\newfont{\mycrnotice}{ptmr8t at 7pt}
\newfont{\myconfname}{ptmri8t at 7pt}
\let\crnotice\mycrnotice%
\let\confname\myconfname%
\usepackage[utf8]{inputenc}
\begin{document}

\title{A Fast Deep Learning Network for Automatic Image Auto-Straightening}

\numberofauthors{5} 

\author{
\alignauthor Ionu\c{t} Mironic\u{a}\\
       \affaddr{Adobe Systems Romania and LAPI, University Politehnica of Bucharest, Romania}\\
       \email{mironica@adobe.com}
\alignauthor Andrei Zugravu\\
       \affaddr{Adobe Systems Romania}\\
       \affaddr{Bucharest, Romania}\\
       \email{zugravu@adobe.com}
}
\maketitle

\begin{abstract}

Rectifying the orientation of images represents a daily task for every photographer. This task may be complicated even for the human eye, especially when the horizon or other horizontal and vertical lines in the image are missing. In this paper we address this problem and propose a new deep learning network specially adapted for image rotation correction: we introduce the rectangle-shaped depthwise convolutions which are specialized in detecting long lines from the image and a new adapted loss function that addresses the problem of orientation errors. 

Compared to other methods that are able to detect rotation errors only on few image categories, like man-made structures, the proposed method can be used on a larger variety of photographs e.g., portraits, landscapes, sport, night photos etc. Moreover, the model is adapted to mobile devices and can be run in real time, both for pictures and for videos. An extensive evaluation of our model on different datasets shows that it remarkably generalizes, not being dependent on any particular type of image. Finally, we significantly outperform the state-of-the-art methods, providing superior results.

\end{abstract}


\section{Introduction}

Image rotation correction represents a tedious task for photographers, and it is one of the most used tools in common Adobe products, such as Photoshop or Lightroom\footnote{http://www.adobe.com}. Photos casually shot by hand-held cameras or phones may appear tilted. The human eyes are sensitive to this even when the rotation angle is small. Modern cameras and phones contain inertial sensors,  but they are only able to correct the orientation of pictures in 90 degree steps, and they are not able to detect the angle errors less than \ang{1}. 

Image angle rotation detection requires high-level scene understanding. Humans use object recognition and contextual scene information to correctly orient images. However, 
even for humans this task is not always easy. In~\cite{Luo2003} a psychophysical study of image orientation perception was developed. The results showed that for typical images, accuracy is close to 98\% only when using all available semantic cues from high-resolution images, and 84\% when using the low-level vision features and coarse semantics from low resolution images. The rapid development in deep learning~\cite{InceptionV3, Densenet, MobilenetV2, Resnet} provides more powerful tools, which are able to learn semantic, high-level, deeper features, may be a good solution for creating an architecture that may achieve similar results to human performance.

In this paper, we present a deep neural network architecture that  automatically correct the image orientation based just on the visual data. To summarize, our main contributions are the following: 
(1) We introduce a new fast deep learning representation for the addressed problem of rotation correction in images and videos,
(2) We demonstrate its generality in terms of applications by applying it to a wide category of images,
(3) We achieve better performance than the state-of-the-art using a fast architecture that can be used in mobile applications. 

The remainder of the paper is organized as follows. In Section~\ref{section:PW} we overview the current state-of-the-art and situate
our work. Section~\ref{section:GA} details the proposed approach. The experimental results are presented in Section~\ref{section:ES} while Section~\ref{section:concl}
concludes the paper and discusses future perspectives.

\begin{figure*}[!ht]
\hspace*{-0.2in}
    \begin{minipage}[t]{1.05\linewidth}
\begin{center}
    \includegraphics[width=\textwidth]{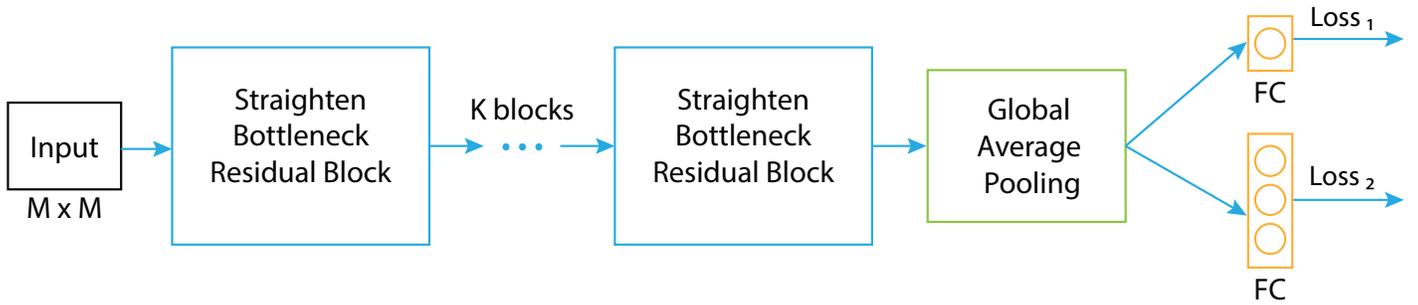}
\end{center}
\caption{General approach of the proposed algorithm.}\label{figProposed}
\end{minipage}
\end{figure*}

\begin{figure}[t]
\begin{minipage}[t]{0.99\linewidth}
\begin{center}
    \includegraphics[width=\textwidth]{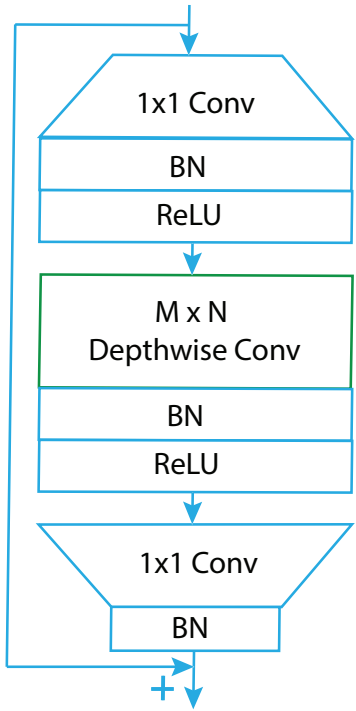}
\end{center}
\caption{Straighten Bottleneck Residual block} \label{figBottleneck}
\end{minipage}
\end{figure}

\section{Related Work}\label{section:PW}

Not much research has been conducted in the field of image rotation detection. Most of the research addresses the subproblem of orientation detection by restricting the angle to be one of \ang{0} (upright orientation), \ang{90}, \ang{180} or \ang{270}. Older methods involve extracting low-level features, such as color moments \cite{vailaya02} or local binary patterns \cite{ciocca13}, and feeding them to a learning algorithm like bayesian learning or logistic regression. Newer methods either use CNNs for the feature extraction part \cite{shima17} or for the whole process \cite{joshi17, swami17}. Some methods work better for certain categories of images, such as images with faces or landscapes \cite{wang03}. Horizon detection \cite{fefilatyev07, lipschutz13} is a special case of image orientation detection, but the horizon is not always visible in many photos.

Lee \textit{et al.} \cite{upright} compute an optimal homography for correcting the rotation. They use camera calibration, which estimates vanishing points and lines as well as camera parameters, in order to approximate part of the parameters required for computing the homography. The rest of the parameters are estimated using an optimization framework.

Fischer \textit{et al.} \cite{alexnet_orientation} use AlexNet to predict the precise rotation angle. They consider 3 difficulty levels: $\pm$\ang{30}, $\pm$\ang{45} and $\pm$\ang{180}. In the latter case, they first predict the orientation using a network with 4 outputs and then predict the exact angle. For precise angle estimation, they use 2 outputs on the final layer to distinguish between positive and negative orientations.

 \hfill \break

\section{Proposed method}\label{section:GA}

We propose a deep neural network which is specially designed for this problem with two novel additions. Firstly, instead of using square convolutions within our network, we use rectangle-shaped convolutions. That is, the convolutions have the shape $M \times N$, where $M$ is less than $N$. These convolutions are better suited for detecting long horizontal lines, which are more important for this type of task. Secondly, we adapt the loss for the network, forcing the network to stronger penalize wrong orientation predictions. 

The detailed network architecture is illustrated in Figure~\ref{figProposed}. It is similar to MobileNetV2~\cite{MobilenetV2} and contains a set of Straighten Bottleneck Residual blocks that are adapted to recognize the rotation angle of the image. The structure of one such block is presented in Figure~\ref{figBottleneck}. There are three convolutional layers in each block. The first layer is a $1 \times 1$ convolution. Its purpose is to expand the number of channels in the data before it goes into the depthwise convolution. This way we can compute more complex features and have finer representations of the data. In contrast, the last layer has the role of reducing the number of channels in the output layer, in order to reduce the number of parameters and computational costs.

The second layer is a rectangular convolution of size $M \times N$, with $M$ less than $N$. This is where most of the computation is done. All convolutions are followed by BN (Batch Normalization) and ReLU activation, except the last convolution in each block, which has a linear activation. After testing, we concluded that a linear activation is preferred over a ReLU activation for the last convolution, just like in the case of MobileNetV2.

We chain $K=16$ of these Straighten Bottleneck Residual blocks. Then, we add a Global Average Pooling layer and connect it with two branches. One branch is a FC (fully connected) layer with one linear neuron for the regression task of predicting the precise rotation angle. The other branch is a FC layer with three neurons for predicting the orientation (one of left, right or no orientation) and softmax as activation function.

Another key point of the network is the adapted loss that is able to reduce the orientation errors. The most disturbing errors of an auto-straighten algorithm are those when the orientation of the prediction is incorrect. For these errors, we propose a new term that penalizes the network when the predicted orientation is wrong. This term is added in the regression loss.

The network loss is comprised of two separate losses for the regression and classification part:
\begin{equation}\label{eqLoss1}
    Loss = w_1 Loss_1 + w_2  Loss_2
\end{equation}
where we set $w_1 = 0.75$ and $w_2=0.25$. $Loss_2$ is the classification loss, represented by cross-entropy, and $Loss_1$ is the regression loss, having the following formula:
\begin{equation}\label{eqRegression}
 Loss_1 = \frac{1}{N} \sum\limits_{c=1}^N  { (y_i - y_{pi}) } ^ 2
          + \frac{\gamma_1}{N} \sum\limits_{c=1}^N  { min(0, y_i  y_{pi}) } ^ 2
          + \gamma_2 {||w||} ^ 2
\end{equation}
where $y_i$ is the ground truth angle and $y_{pi}$ is the predicted angle. $\gamma_1$ and $\gamma_2$ are the weights of the corresponding terms and $N$ is the batch size.

The regression loss consists of $3$ terms. The first term represents the mean squared error, the classic regression loss, measuring the differences between the predicted angle and the true rotation angle. The second term represents the novel orientation loss, penalizing the orientation errors that are most disturbing. When $y_i$ and $y_{pi}$ have different signs, meaning different orientations, their product will be less than 0 and, when squared, the loss will increase. When they have the same sign, their product will be positive and the loss will not increase. The final term is the regularization component that penalizes the weights of the coefficients.

\section{Experiments}\label{section:ES}

\subsection{Dataset and evaluation}
We demonstrate the advantages and generality of our algorithm on two different datasets. The total size of the first dataset is of $508,859$ images. The initial images are downloaded from the Pixabay\footnote{http://www.pixabay.com} web platform. We choose this source because most of the photographs are uploaded by professional photographers and almost all of them have already a correct rotation angle. Also, the images that needed angle corrections were manually corrected by annotators using Lightroom software\footnote{http://www.adobe.com}. The images are augmented to various angles starting from \ang{-12} to  \ang{12}. This dataset is used to tune the model parameters and to compare our proposed method to other baseline deep learning architectures.

The second dataset contains a larger variety of images with angles starting from \ang{-25} to \ang{25}. The total number of photographs is approximately $1,500,000$. Compared to almost all the previously proposed methods that are trying to correct only one type of images (e.g., Google street images, images with buildings), the dataset contains images from various categories: city photographs, landscapes, portraits, sports  and night images. The main purpose of this dataset is to create a comprehensive experiment that can provide a relevant comparison  with other state-of-the-art proposals. Each dataset is balanced, i.e., each of the orientation classes have equal number of images.

Performance is assessed with two metrics. The first one is MAE (Mean Absolute Error) which represents the mean of absolute angle errors. The second is a measure that computes the percentage of predictions that have an angle error less than \ang{1}, and it will be referred in the experimental section as accuracy.

\subsection{Results on evaluation dataset}

\begin{table}[!t]
\centering \caption{Comparison with state-of-the-art transfer learning  in terms of accuracy and mean absolute error on the first dataset.}
\label{tabAutoStraightenEvaluation}
\scalebox{0.95}{
\begin{tabular}{|p{4cm}|p{1.8cm}|p{1.6cm}|} \hline
\textbf{Architecture}                           & \textbf{Accuracy}   & \textbf{MAE}   \\ \hline
Mobilenet V2~\cite{MobilenetV2}      & 75.07\%   & 1.04   \\ \hline
Resnet 50~\cite{Resnet}                    & 67.75\%  & 1.98   \\ \hline
DenseNet 121~\cite{Densenet}         & 77.55\%  & 1.12   \\ \hline
InceptionV3~\cite{InceptionV3}       & 57.26\%  & 1.98   \\ \hline
Proposed & 98.36\% & 0.21 \\ \hline
\end{tabular}}
\end{table}

\textbf{Training setup.}  We train our models using Keras~\cite{chollet2015keras}. We use the standard RMSProp optimizer with both decay and momentum set to $0.9$.
We set batch normalization after every layer, and the learning rate is equal to $10^{-3}$, while number of epochs was equal to $50$. We decrease the learning rate on each epoch using a standard weight decay that is set to $l/N$, where $l$ represents the learning rate and $N$ is the number of epochs. To prevent overfitting, we perform image augmentation, i.e., we apply random transformations to input samples during network training on the fly. We use brightness adjustment in the range $\mathopen[-0.1,0.1\mathclose]$ and gamma adjustment with $\gamma \in [0.7,1.4\mathclose]$. We did not apply cropping because it often removes important semantic cues from images. 

\textbf{Results.}  In this experiment we want to compare our proposed architecture with a classical transfer learning approach that may use one of the state-of-the-art pretrained deep learning architectures: Inception-V3~\cite{InceptionV3} , Densenet-121~\cite{Densenet}, Mobilenet~\cite{MobilenetV2} and Resnet-50~\cite{Resnet}. In order to increase the convergence speed, we initialize our weights with the pretrained weights from the Imagenet dataset~\cite{ImageNet}. The goal of this experiment is to demonstrate that the proposed architecture  has a great impact on the accuracy of auto-straighten problem. All the experiments are performed on the evaluation dataset. For each state-of-the-art deep learning network we removed the last layers and added a global average pooling and a dense layer with linear activation. As loss function we used the mean square error. The results are provided in Table~\ref{tabAutoStraightenEvaluation} and show that we have obtained better results with significant improvements: the accuracy goes up from $77.55\%$ to $98.36\%$ and the MAE decreases from $1.04$ to $0.21$.

We conclude that our explicit model is very effective for the automatic rotation correction problem.

\subsection{Comparison with State-of-the-art}

The experiment provided in this section is performed on the second dataset. For our model we used the same parameters and methodology from the previous experiment.

\begin{figure*}[!ht]
    \begin{minipage}[t]{\linewidth}
\begin{center}
    \includegraphics[width=\textwidth]{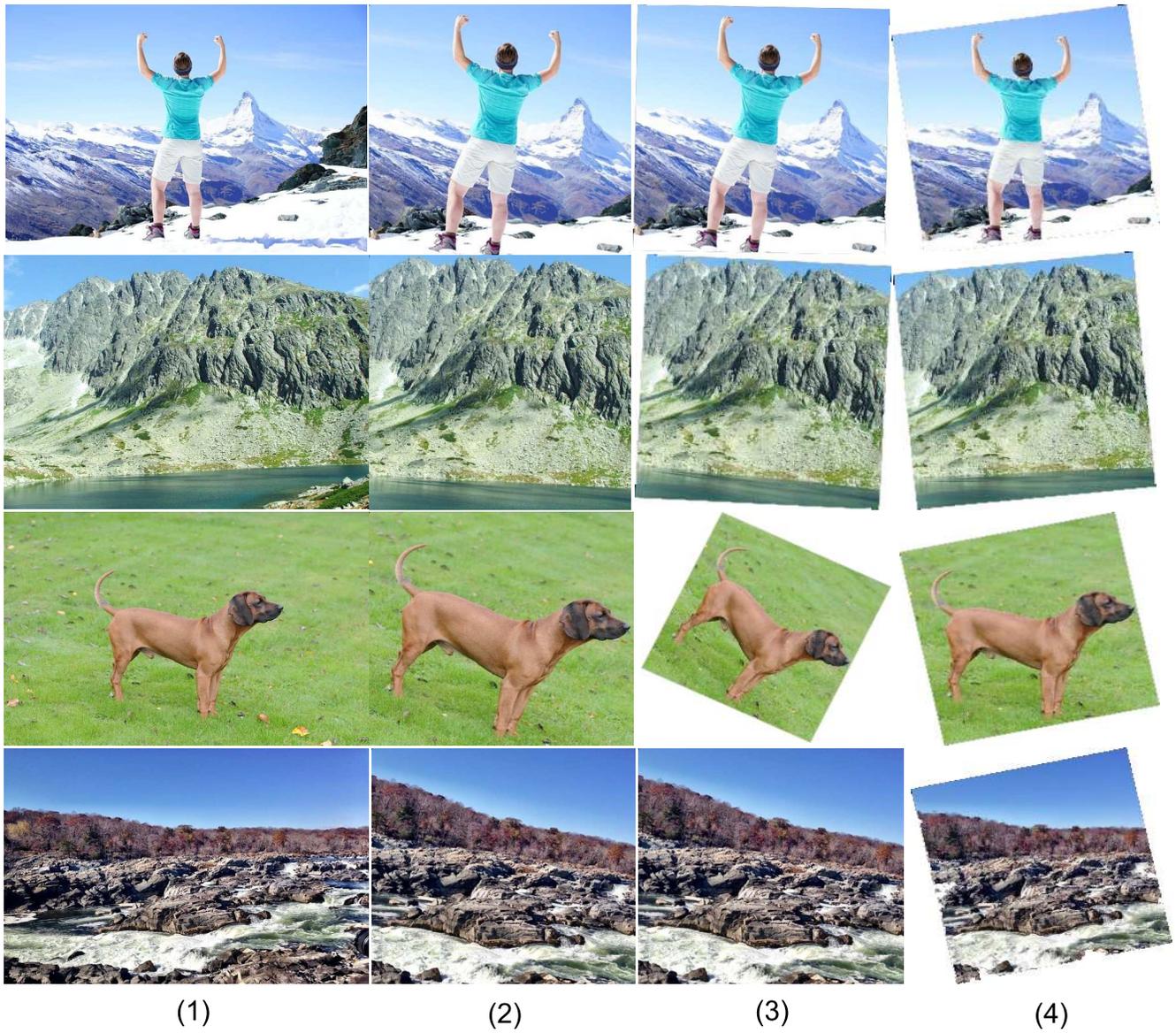}
\end{center}
\caption{Qualitative results of our method. First row shows ground truth input images. Second row images provides a crop from the rotated ground truth image. The third row contains the angle estimation correction provided by~\cite{upright} while the fourth column  shows images rotated according to the predicted orientation label of our algorithm} \label{figExamples}
\end{minipage}
\end{figure*}

In this section, we compare our approach against state-of-the-art results from the literature; in particular with~\cite{alexnet_orientation}, using our implementation of the paper,  and~\cite{upright}, using the original code provided by the authors. The  quantitative results of our model compared to ~\cite{alexnet_orientation, upright} are presented in Table~\ref{tabAutoStraightenSOA}. Both methods obtain reasonable results, namely $3.15$ and $6.40$ MAE values. 
\cite{upright}  fails to converge on many cases since multiple images in our
test set do not contain edges on which the technique relies on. On the other side,the network from ~\cite{alexnet_orientation} is not capable to generalize many complex cases from the dataset. Our model achieved an impressive accuracy of $92.46\%$ and a MAE value of $0.62$. This shows the remarkable generalization capability of our model which detects correct orientation angle of a large variety of images outside the training dataset. Furthermore, our obtained results are much better than other state-of-the-art algorithms. Figure~\ref{figExamples} shows the qualitative results obtained with our model for some of the challenging images from different categories. We have also presented a comparison with the prediction provided by~\cite{upright}.

To conclude, using the proposed model significantly improves the accuracy and the MAE, yielding much better results than the state-of-the-art algorithms.

\subsection{Automatic video rotation correction}

A single forward pass of our networks takes 26 milliseconds at an input resolution of $848\times480$ pixels on an NVIDIA RTX 2080 Ti GPU, which enables us to use it in real-time applications like video stabilization. The network successfully generalizes to videos captured with the webcam or phone camera. The fact that the network works with videos proves that it does not make use of interpolation artifacts which appear when artificially applying rotations. 

\begin{table}[!t]
\centering \caption{Comparison with state-of-the-art methods  in terms of accuracy and mean absolute error on the second dataset.}
\label{tabAutoStraightenSOA}
\scalebox{0.95}{
\begin{tabular}{|p{4cm}|p{1.8cm}|p{1.6cm}|} \hline
\textbf{Method}     & \textbf{Accuracy}      & \textbf{MAE}   \\ \hline
Fischer at al.~\cite{alexnet_orientation}     & 57.17\%   & 3.15     \\ \hline
Lee at al.~\cite{upright}                               & 39.53\%   & 6.40   \\ \hline
Proposed                                                       & 92.46\%   &  0.62         \\ \hline
\end{tabular}}
\end{table}

\section{Conclusions}\label{section:concl}
 
In this paper we addressed the problem of rotation correction in images and videos. We proposed a novel deep learning architecture that is able to recognize the angle rotation error of an image with high accuracy. 
We demonstrated that our proposed method is highly efficient: (1) we showed significant improvements on a wide
variety of images, ranging from landscapes, to city photos, and to portraits and (2) we significantly improved the results over several state-of-the-art algorithms, providing superior results. Also, the network runs in realtime, which allows us to apply it to live video streams. 

In future work we plan to generalize the architecture of the model to be able to detect other perturbations of the image, like perspective errors or camera calibration errors.

\bibliographystyle{ieeetr}
\bibliography{sigproc}


\end{document}